\pdfoutput=1
\documentclass[11pt]{article}

\usepackage[final]{acl}
\usepackage{bbm}
\usepackage{amssymb, amsmath,fge}
\usepackage{booktabs}
\usepackage{times}
\usepackage{latexsym}
\usepackage[most]{tcolorbox}
\usepackage{multirow}
\usepackage{enumitem}
\usepackage{subcaption}
\usepackage[T1]{fontenc}
\usepackage[utf8]{inputenc}

\usepackage{microtype}

\usepackage{inconsolata}

\usepackage{graphicx}
\usepackage[bidi=default, english]{babel}
\usepackage[LAE, T1]{fontenc}
\usepackage{titlesec}

\titlespacing{\section}{0.2pt}{\parskip}{-\parskip}
\titlespacing{\subsection}{0.2pt}{\parskip}{-\parskip}
\titlespacing{\subsubsection}{0.2pt}{\parskip}{-\parskip}
\title{MizanQA: Benchmarking Large Language Models on Moroccan Legal Question Answering\thanks{Preprint version.}}

\author{Adil Bahaj \\
  Mohammed 6 Polytechnic University\\
  \\\And
  Mounir Ghogho \\
  Mohammed 6 Polytechnic University \\
  \\}

\begin{document}
\maketitle
\vspace{-20cm}
\begin{abstract}
The rapid advancement of large language models (LLMs) has significantly propelled progress in natural language processing (NLP). However, their effectiveness in specialized, low-resource domains—such as Arabic legal contexts—remains limited. This paper introduces MizanQA (\foreignlanguage{arabic}{ميزان} pronounced Mizan, meaning "scale" in Arabic, a universal symbol of justice), a benchmark designed to evaluate LLMs on Moroccan legal question answering (QA) tasks, characterised by rich linguistic and legal complexity. The dataset draws on Modern Standard Arabic, Islamic Maliki jurisprudence, Moroccan customary law, and French legal influences. Comprising over 1,700 multiple-choice questions, including multi-answer formats, MizanQA captures the nuances of authentic legal reasoning. Benchmarking experiments with multilingual and Arabic-focused LLMs reveal substantial performance gaps, highlighting the need for tailored evaluation metrics and culturally grounded, domain-specific LLM development.
\end{abstract}
\section{Introduction}

The proliferation of large language models (LLMs) has instigated transformative advancements across numerous disciplines through their enhanced natural language understanding and generation capacities. Nevertheless, the applicability and efficacy of these models within specialized domains, such as legal contexts, particularly in low- and medium-resource languages like Arabic, continue to constitute active areas of scholarly inquiry. This paper delineates ongoing research focused on evaluating the proficiency of large language models in comprehending and processing Arabic legal corpora for the Moroccan legal system. 

Moroccan legal language furthers the complications that Arabic represents LLMs \cite{bayan2025can, daoud2025medarabiq}. Moroccan law is written in Modern Standard Arabic, but it is infused with local legal idioms and cultural references. Scholars observe that Moroccan legal texts are shaped by a blend of influences – Islamic Maliki jurisprudence, Moroccan customary law, and remnants of French and international law – which introduces “cultural specificities inherent to legal terminology” \cite{mellouki2025issues}. In practice, this means that statutes and regulations may use archaic or region-specific expressions that do not appear in standard Arabic corpora. For NLP models, this mix of formal Arabic syntax and Moroccan-specific terms makes understanding the content especially challenging. Accurate legal QA in this context requires handling precise legal phrasing while recognizing concepts that are unique to Morocco’s legal system.
This study introduces \textbf{MizanQA}, a legal benchmark designed to evaluate the performance of existing large language models (LLMs) in answering legal questions within the Moroccan legal framework. The benchmark comprises over 1,700 multiple-choice question-answer pairs, encompassing a range of complexities—from questions testing basic legal knowledge to those requiring detailed understanding of specific legal articles and reasoning abilities. A distinctive feature of MizanQA is the inclusion of questions that necessitate the selection of multiple correct options, thereby increasing the overall difficulty of the task.

In summary, this paper makes the following key contributions:
\begin{itemize}[leftmargin=0.1cm]
    \setlength\itemsep{0.005cm}
    \item The release of \textbf{MizanQA} \footnote{\url{https://huggingface.co/datasets/adlbh/MizanQA-v0}}, a high-quality, manually curated dataset of Arabic multiple-choice questions and answers covering Moroccan law.
    \item A detailed evaluation of leading LLMs (both multilingual and Arabic-centric) on the MizanQA benchmark.
    \item A proposal of new evaluation metrics to mùeasure response accuracy and confidence calibration for multiple-choice QA that handle questions with more than one correct option.
\end{itemize}

\section{Related Work}
    The success of multilingual LLMs (e.g., GPT \cite{openai2024gpt4technicalreport}, Gemini \cite{yang2024advancing, team2023gemini}) has spurred the development of native Arabic models such as ALLAM \cite{bari2024allam} and JAIS \cite{sengupta2023jais}. Despite this progress, these models exhibit notable domain-specific knowledge gaps that limit their applicability \cite{bayan2025can, daoud2025medarabiq}. Existing legal benchmarks, while diverse in task complexity, are predominantly in English and focused on English-speaking jurisdictions \cite{fei2024lawbench, hijazi2024arablegaleval, guha2023legalbench, pipitone2024legalbench, li2024legalagentbench, dahl2024large}, with limited exceptions like Chinese \cite{fei2024lawbench, li2024legalagentbench} and Saudi Arabic datasets \cite{hijazi2024arablegaleval}. To date, only one Arabic legal benchmark exists \cite{hijazi2024arablegaleval}, primarily featuring translated content and Saudi law. This work introduces the first legal QA dataset for Moroccan law, addressing its unique linguistic and cultural complexity. Unlike prior benchmarks, which assume single-answer MCQs with uniform option sets, Moroccan legal assessments often require selecting multiple correct options from variable sets—prompting the development of new evaluation metrics tailored to this context.  
\section{MizanQA Dataset}
\subsection{Data sources}
    MizanQA is extracted based on publicly available Moroccan law MCQ banks and exams.
\subsection{Construction Process}
    The construction process of the dataset went through multiple phases with hybrid manual and automated steps.
    \begin{itemize}[leftmargin=0.1cm]
        \item \textbf{Step 1: collection}
        a set of publicly available sources of Moroccan law MCQs is extracted 
        \item \textbf{Step 2: temporal curation}
        The collected documents were curated by a legal expert to sift out any documents that use outdated legislation.
        \item \textbf{Step 3: organisation}
        The MCQs were grouped into image batches to facilitate automated question extraction. For structured documents with a fixed number of self-contained MCQs per page, this process was automated. In contrast, irregular documents—where MCQs spanned multiple pages or answers were consolidated at the end—required manual organization. This involved capturing screenshots to ensure each page was self-contained, with complete questions, options, and answers, which were then converted into images.
        \item \textbf{Step 4: Extraction}
        The images containing batches of MCQs produced in the previous step are fed to a multimodal LLM (i.e. Gemini-2.0-flash in our case) to extract MCQs in a standardised format.
        \item \textbf{Step 5: verification}
        The extracted MCQs in the previous step are verified manually. The curators follow a set of verification guidelines (appendix \ref{sec:verif_guid}) to ensure that the extracted questions are similar to the original ones.
        \item \textbf{Step 6: categorisation}
        Depending on the original documents, MCQs are categorised manually based on the set of legislation they represent (e.g. Criminal law, constitution, etc). This is followed by normalisation of the categories to remove any redundancy.
    \end{itemize}    

\section{Benchmarking Study}
\begin{table}[h]
    \centering
    
    \begin{subtable}[t]{0.48\textwidth}
        \centering
        \setlength{\tabcolsep}{1.7pt}
        \scalebox{0.8}{
        \begin{tabular}{lrrrr}
        \toprule
         Model& PM(1) $\uparrow$& PM(0.5)$\uparrow$ & F1(1) $\uparrow$& F1(2) $\uparrow$\\
        \midrule
        Allam-2 (7b) & 26.88 & 34.04 & 43.07 & 39.93 \\
        Gemini-1.5-flash & 35.90 & 44.23 & 53.30 & 48.93 \\
        Gemini-2.0-flash & 53.57 & 58.34 & 64.84 & 62.16 \\
        Llama-3.3 (70b) & 46.78 & 50.73 & 59.21 & 56.18 \\
        Llama-4-maverick (17b) & 49.97 & 55.53 & 64.90 & 61.29 \\
        Llama-4-scout (17b) & 44.06 & 49.01 & 59.51 & 55.60 \\
        \bottomrule
    \end{tabular}}
    \caption{$\text{F1}(\alpha)$ refers to the F1-Like metric in equation \ref{eq:f1_meas}. and $\text{PM}(\beta)$ refers to the measure in equation \ref{eq:pmpa_meas}.}
    \label{tab:table1_a}
    \end{subtable}
    \begin{subtable}[t]{0.48\textwidth}
        \centering
        \setlength{\tabcolsep}{1.7pt}
        \scalebox{0.8}{
        \begin{tabular}{lrrr}
        \toprule
        Model & ACC $\uparrow$ & $\text{ECE}_{\text{opt}}\downarrow $& $\text{ECE}_{\text{set}}\downarrow$ \\
        \midrule
        Allam-2 (7b) & 15.32 & 28.42 & 51.43 \\
        Gemini-1.5-flash & 24.26 & 34.77 & 48.52 \\
        Gemini-2.0-flash & 42.11 & 28.15 & 41.16 \\
        Llama-3.3 (70b) & 33.28 & 35.27 & 59.40 \\
        Llama-4-maverick (17b) & 36.83 & 17.64 & 29.10 \\
        Llama-4-scout (17b) & 31.27 & 36.99 & 61.78 \\
        \bottomrule
    \end{tabular}}
    \caption{$\text{ACC}$, $\text{ECE}_{\text{opt}}$ and $\text{ECE}_{\text{set}}$ refers to the measures in equations \ref{eq:acc_meas}, and the options and set variants of equation \ref{eq:ECE} respectively.}
    \label{tab:table1_b}
    \end{subtable}
    \caption{Evaluation results of various models on MizanQA.}
    \label{tab:res_table}
    \vspace{-15pt}
\end{table}
\subsection{Evaluation metrics}
    \paragraph{Accuracy Measures}
    We found that most MCQs from Moroccan sources have multiple options. An answer is considered correct only if all the right options are chosen. We didn't find any instances in LLM QA literature with this setting. Consequently, we created different performance metrics to evaluate LLMs on this task. Let $\mathcal{Q}={(Q_i, O_i, C_i)}_i$ be the set of questions $Q_i$, their corresponding options $Q_i$ and the correct options $C_i$. Let $\mathbf{P}(Q_i, O_i)$ be a prompt parameterised by question $Q_i$ and its corresponding options $O_i$ and let $S_i=\text{LLM}(\mathbf{P}(Q_i, O_i))$ be the set of options predicted by an LLM to be correct for question $Q_i$. $S_i=\{(o_j, p_j)\}_j$ is composed out of tuples $(o_j, p_j)$, where $o_j\in O_i$ is an option selected by the LLM and $p_j\in [0, 1]$ is the LLMs corresponding confidence that option $j$ is the right option. We define strict accuracy as:
    \begin{equation}
        \label{eq:acc_meas}
        \text{ACC}=\frac{1}{|\mathcal{Q}|}\sum_{i}^{|\mathcal{Q}|} \mathbbm{1}_{[S_i\setminus C_i=C_i\setminus S_i=\emptyset]}
    \end{equation}
    $\mathbbm{1}_{[A]}$ Is the indicator function, which equals 1 if $A$ is true and 0 otherwise. $\text{ACC}$ rewards only perfectly correct answers. Additionally, to reward partial correctness while penalising incorrect selections, we propose a metric inspired by the F1 metric \cite{sitarz2022extending}:

    \begin{equation}
        \label{eq:f1_meas}
        \text{F1-like}_\alpha=\frac{1}{|\mathcal{Q}|}\sum_{i}^{|\mathcal{Q}|}\frac{2P_iR_i}{P_i+R_i}
    \end{equation}
    where $R_i=\frac{TP_i}{TP_i+FN_i}$ is equivalent to recall and $P_i=\frac{TP_i}{TP_i+\alpha\cdot FP_i}$ is equivalent to precision, such that $TP_i=|C_i\cap S_i|$, $FP_i=|S_i\setminus C_i|$ and $FN_i=|C_i\setminus S_i|$ are true positives (correct answers selected), false positives (wrong answers selected) and false negative (missed correct answers), respectively. $\alpha\ge 1$ increases the penalty for wrong choices. We also propose Partial Match Penalized Accuracy (PMPA):
    \begin{equation}
        \label{eq:pmpa_meas}
        \scalebox{0.9}{
        $\text{PMPA}_\beta=\frac{1}{|\mathcal{Q}|}\sum_{i}^{|\mathcal{Q}|}\max\left(0, \min\left(1, \frac{TP_i-\beta\cdot FP_i}{|C|}\right)\right)$}
    \end{equation}
    where $\beta\in[0,1]$ is a penalty factor for incorrect answers. The F1-like score and the PMPA score have a similar objective, but the PMPA score is more advantageous in cases where the number of correct options varies significantly. This is particularly important since the number of options per question in our dataset varies from 2 to 16.
    \paragraph{Confidence calibration measures}
    A model exhibits well-calibrated uncertainty when its predicted probabilities are congruent with observed empirical frequencies; specifically, events assigned a probability $p$ occur with a relative frequency of $p$ in empirical validation. Following \cite{naeini2015obtaining}, we estimate Expected Calibration Error (ECE) by binning the maximum output probability of each of $N$ samples into $M$ equally-spaced bins $B = \{B_m\}_{m=1}^{M}$ w.r.t. the prediction confidence estimated for each sample. The empirical ECE estimator is given by,
    
    \begin{equation}
        \label{eq:ECE}
        \text{ECE}= \sum_{m=1}^{M}\frac{|B_j|}{N}|\text{conf}(B_j)- \text{acc}(B_j)|
    \end{equation}
    We use this measure in two settings: a) the Per-Option Calibration and b) Set-Level Calibration.
    \begin{itemize}[leftmargin=0.1cm]
        \item \textbf{Per-Option Calibration Setting}: Let $\mathcal{D}_{\text{opt}}=\{(y_{i, j}, p_{i, j})\}$ such that $i$ is the index of examples and $j$ is the index of options (i.e. $j$th predicted option of the $i$th), which that Let $y_{i, j} = \mathbbm{1}_{[o_{i, j}\in O_i]}$.
        \begin{itemize}[leftmargin=0.3cm]
            \item The empirical accuracy in bin $B_{m}$ is:
            \begin{equation}
                \text{acc}
                (B_m)=\frac{1}{|B_m|}\sum_{(y, p)\in B_m} \mathbbm{1}_{[y=1]}
            \end{equation}
            \item The average predicted confidence is:
            \begin{equation}
                \text{conf}(B_m)=\frac{1}{|Bm|}\sum_{(y, p)\in B_m}p
            \end{equation}
            \item Number of examples $N$:
            $N=|\mathcal{D}_{\text{opt}}|$
        \end{itemize}
        \item \textbf{Set-Level Calibration}: let $\mathcal{D}_{\text{set}}=\{(z_i, q_i)\}_i$ such that $z_i=\mathbbm{1}_{[O_i=C_i]}$ is an indicator which equals 1 if and only if the predicted set exactly matches the ground truth. We define the set-level confidence as the product of confidences for the selected options: $q_i=\prod_{(o_j, p_j) \in S_i} p_{j}$. This can be interpreted as the model's implicit confidence that each selected option is correct, under independence. After bining the pairs $(z_i, q^i)$ the following metrics can be calculated :
        \begin{itemize}[leftmargin=0.3cm]
            \item Empirical accuracy in each bin ($\text{acc}(B_m)$):
            \begin{equation}
                \text{acc}(B_m)=\frac{1}{B_m}\sum_{(z_i, q_i)\in B_m} z_i
            \end{equation}
            \item Average predicted joint confidence ($\text{conf}(B_m)$):
            \begin{equation}
                \text{conf}(B_m)=\frac{1}{B_m}\sum_{(z_i, q_i)\in B_m}q_i
            \end{equation}
            \item Number of examples $N$: $N=|\mathcal{D}_{\text{set}}|$
        \end{itemize}
    \end{itemize}
    Practically, the Per-Option Calibration Setting ($\text{ECE}_{\text{opt}}$)  and the Set-Level Calibration error ($\text{ECE}_{\text{set}}$) are obtained by replacing their respective expressions of $\text{conf}(B_m)$, $\text{acc}(B_m)$ and $N$ in equation \ref{eq:ECE}.
\subsection{Baselines}
    We evaluated various multilingual and specialised Arabic LLMs on MizanQA. These models have varying levels of complexity (i.e. number of parameters, support for reasoning etc). We evaluated the following models: Allam-2 (7b) \cite{bari2024allam}, Gemini-1.5-flash \cite{yang2024advancing, team2023gemini}, Gemini-2.0-flash \cite{yang2024advancing, team2023gemini}, Llama-3.3 (70b) \cite{grattafiori2024llama}, Llama-4-maverick (17b) \cite{metaLlamaHerd}, and Llama-4-scout (17b) \cite{metaLlamaHerd}.

\subsection{Experimental Setting}
    The questions are given to the LLMs using a prompt template where the question and options are replaced, and the LLM is tasked with generating a list containing the right options, in addition to its confidence in each option being right. The prompt template is described in Appendix \ref{sec:bench_app}.
    
\subsection{Results}
Table 1 reports the performance of various LLMs on the MizanQA benchmark, with Gemini models generally outperforming Llama and Allam-2 across most metrics. Gemini-2.0-flash achieves the highest scores in PMPA, F1-Like, and ACC, while Llama-4-maverick shows superior calibration with the lowest ECE. A decline in performance is observed as penalties for incorrect answers increase. These results highlight MizanQA’s difficulty and reveal notable regional knowledge gaps in both open and closed LLMs.

% \begin{table}
%     \centering
%     \setlength{\tabcolsep}{1.7pt}
%     \scalebox{0.8}{
%     \begin{tabular}{lrrrrrrr}
%     \hline
%     Model & PMPA(1)$\uparrow$ & PMPA(0.5)$\uparrow$ & F1-Like(1) $\uparrow$& F1-Like(2) $\uparrow$& ACC $\uparrow$ & $\text{ECE}_{\text{opt}}\downarrow $& $\text{ECE}_{\text{set}}\downarrow$ \\
%     \midrule
%     Allam-2 (7b) & 26.88 & 34.04 & 43.07 & 39.93 & 15.32 & 28.42 & 51.43 \\
%     Gemini-1.5-flash & 35.90 & 44.23 & 53.30 & 48.93 & 24.26 & 34.77 & 48.52 \\
%     Gemini-2.0-flash & 53.57 & 58.34 & 64.84 & 62.16 & 42.11 & 28.15 & 41.16 \\
%     Llama-3.3 (70b) & 46.78 & 50.73 & 59.21 & 56.18 & 33.28 & 35.27 & 59.40 \\
%     Llama-4-maverick (17b) & 49.97 & 55.53 & 64.90 & 61.29 & 36.83 & 17.64 & 29.10 \\
%     Llama-4-scout (17b) & 44.06 & 49.01 & 59.51 & 55.60 & 31.27 & 36.99 & 61.78 \\
%     \bottomrule
%     \end{tabular}}
%     \caption{Evaluation results of various models on MizanQA.}
%     \label{tab:res_table}
%     \vspace{-20pt}
% \end{table}

\subsubsection{Performance by category}
Appendix \ref{tab:res_table_by_cat} presents a comparative analysis of LLM performance across categories of Moroccan law, revealing a general improvement from Allam-2 (7b) to Gemini-2.0-flash, with Gemini models outperforming the Llama series. Higher accuracy in the Law of Obligations and Contracts and the Moroccan Constitution suggests these domains may be less complex for LLMs, potentially due to their alignment with international legal standards. In contrast, lower performance in the Family Code and Criminal Law reflects challenges associated with their integration of Islamic jurisprudence and human rights frameworks. Calibration errors vary across models and categories, indicating inconsistencies between model confidence and predictive accuracy.
\section{Conclusion}

This paper introduces MizanQA, the first benchmark tailored to assess legal reasoning in large language models within the context of Moroccan law. Comprising over 1,700 expert-validated multiple-choice questions derived from authentic legal texts, MizanQA reflects the linguistic and conceptual complexity of Moroccan legal discourse. Evaluation of leading LLMs reveals baseline competence but highlights limitations in handling culturally specific terminology, complex reasoning, and multi-answer formats. The findings emphasize the need for domain-specific benchmarks that capture the legal and linguistic diversity of low-resource contexts, promoting equitable development and assessment of legal AI systems.

\section*{Limitations}
    This work represents an initial step towards creating a universal benchmark and legal LLMs for all Arab countries. We chose Moroccan law to be our initial exploration because of its inherent complexity and deviation from the laws of other Arab countries in terms of legislation and wording. The limitations of the dataset can be summarized as follows:
    \begin{itemize}[leftmargin=0.1cm]
        \item \textbf{Coverage Bias}: The dataset does not comprehensively represent Moroccan law, particularly in undercodified areas, region-specific legal practices, and recent legislative updates. Furthermore, it lacks coverage of legal systems from other Arab countries.
        \item \textbf{Lack of Real-World Complexity}: Although it includes reasoning-based and multi-answer questions, the dataset may still oversimplify the complex, interpretive nature of legal reasoning encountered in actual legal practice.
        \item \textbf{Overreliance on Multiple Choice Format}: While useful for benchmarking, multiple-choice formats may not fully reflect how legal professionals reason, argue, or interpret texts.
    \end{itemize}
\section*{Ethics Statement}
This work presents MizanQA, a research-oriented legal QA benchmark based on Moroccan law, constructed from official public-domain sources while excluding sensitive data. All QA pairs were manually verified for accuracy and relevance, with attention to minimizing bias. The benchmark is intended solely for evaluation and not as a substitute for legal advice. Emphasizing the ethical implications of legal AI, the study advocates for transparency, fairness, and human oversight. No human subjects were involved, and no ethical approval was required.
% \subsection{References}

\bibliography{custom}

\newpage

\appendix

\section{Dataset Description}
\label{app:data_desc}
Table \ref{tab:gen_stats} summarises different statistics of MizanQA. The dataset contains a varying number of options and correct answers, which increases the complexity of the benchmark. Table \ref{tab:category_distribution} lists the number of questions per legal topic category. Table \ref{tab:example_mizan} gives an example of a question present in MizanQA. The dataset is publicly available at \url{https://huggingface.co/datasets/adlbh/MizanQA-v0}.

\begin{table*}[htbp]
    \centering
    \begin{tabular}{c|c}
        \hline
         Statistic& Values \\
         \hline
         Number of questions&1776\\
         Number of categories&14\\
         Number of options per question& min: 2, max: 12\\
         Number of words per question & min: 1, max: 63\\
         Number of correct options per question & min: 1, max: 10\\
         Number of words per options& min: 1, max: 71\\
         \hline
    \end{tabular}
    \caption{General statistics of MizanQA. min and max signify the range of values that a statistic has in the MizanQA.}
    \label{tab:gen_stats}
\end{table*}
\begin{table*}[htbp]
    \centering
    \begin{tabular}{llr}
    \hline
    Category (EN) & Category (EN) & Count \\
    \hline
    Civil Procedure & \foreignlanguage{arabic}{المسطرة المدنية} & 460 \\
    Criminal Law & \foreignlanguage{arabic}{القانون الجنائي} & 847 \\
    Exam & \foreignlanguage{arabic}{الامتحانات} & 131 \\
    Family Code & \foreignlanguage{arabic}{مدونة الأسرة} & 38 \\
    Family Law & \foreignlanguage{arabic}{المادة الأسرية} & 66 \\
    Law of Obligations and Contracts & \foreignlanguage{arabic}{قانون الالتزامات والعقود} & 37 \\
    The Judicial System of the Kingdom & \foreignlanguage{arabic}{التنظيم القضائي للمملكة} & 88 \\
    The Justice Sector & \foreignlanguage{arabic}{قطاع العدل} & 39 \\
    The Moroccan Constitution & \foreignlanguage{arabic}{الدستور المغربي} & 70 \\
    \hline
    \end{tabular}
    \caption{Distribution of topic categories in MizanQA.}
    \label{tab:category_distribution}
\end{table*}
    \begin{table*}[]
        \centering
        \begin{tabular}{c|p{16em}|p{16em}}
             &Arabic & English Translation \\
             \hline
            Question& \foreignlanguage{arabic}{إذا نسب لباشا أو خليفة أول لعامل، أو رئيس دائرة أو قائد أو لضابط شرطة قضائية غير المشار إليهم سابقا، ارتكابهم لجناية أو جنحة اثناء مزاولة مهامهم، فإن}&If it is alleged that a Pasha, a first deputy to a governor, a head of a department, a commander, or a judicial police officer other than those previously mentioned, has committed a felony or misdemeanor while performing their duties, then\\
            Options & 'A': \foreignlanguage{arabic}{'الرئيس الأول لمحكمة الاستئناف المعروضة عليه القضية من طرف الوكيل العام للملك إذا قرر إجراء بحث فإنه يعين مستشارا مكلفا بالتحقيق بمحكمته'}, 'B': \foreignlanguage{arabic}{'إذا تعلق الأمر بجناية فإن المستشار المكلف بالتحقيق يصدر أمرا بإحالة القضية إلى غرفة الجنايات'}, 'C': \foreignlanguage{arabic}{'إذا تعلق الأمر بجنحة، فإنه يحيل القضية إلى محكمة ابتدائية غير التي يزاول المتهم فيها مهامه'}, 'D': \foreignlanguage{arabic}{'يرجع الاختصاص إلى محكمة النقض إذا كان ضابط الشرطة القضائية مؤهلا لمباشرة وظيفته في مجموع تراب المملكة'}, 'E': \foreignlanguage{arabic}{'يمكن للطرف المدني التدخل لدى هيئة الحكم'}, 'F': \foreignlanguage{arabic}{'جميع الأجوبة صحيحة'} & 'A': The first president of the Court of Appeal to whom the case is referred by the Public Prosecutor, if he decides to conduct an investigation, shall appoint an advisor in charge of the investigation in his court., 'B': If it is a felony, the investigating advisor issues an order referring the case to the criminal chamber., 'C': If it is a misdemeanor, he refers the case to a court of first instance other than the one in which the accused performs his duties., 'D': Jurisdiction reverts to the Court of Cassation if the judicial police officer is qualified to perform his duties throughout the Kingdom., 'E': The civil party may intervene before the arbitral tribunal., 'F': The answers are correct\\
        Answer &F&F\\
        \end{tabular}
        \caption{An example of a Question and its corresponding answer in MizanQA.}
        \label{tab:example_mizan}
    \end{table*}
\section{Construction process}
    The construction process of MizanQA is semi-automated. It is composed out of multiple steps, some of which are automated while others require human intervention. We observed that a significant number of documents are based on outdated legislation; consequently, to remove these documents, \textbf{Step 2} was included. The motivation behind \textbf{steps 3} and \textbf{4} is the problems faced by annotators when copying in pasting Arabic text from PDFs. The vast majority of documents, when copied and pasted, produce unreadable information. Consequently, optical character recognition (OCR) was essential to automate the extraction. Although the automated extraction is highly accurate, the LLM produces some mistakes (e.g. not listing all the right answers, etc). To eliminate these issues \textbf{step 5} is conducted for manual verification. In the last step, MCQs are categorised depending on the original documents from which they were extracted, and the categories are normalised to remove any redundancies made by the annotators. In what follows, we give more details about the construction process.

\subsection{Step 1: Collection}
\begin{figure}[htbp]
        \centering
        \includegraphics[scale=0.3]{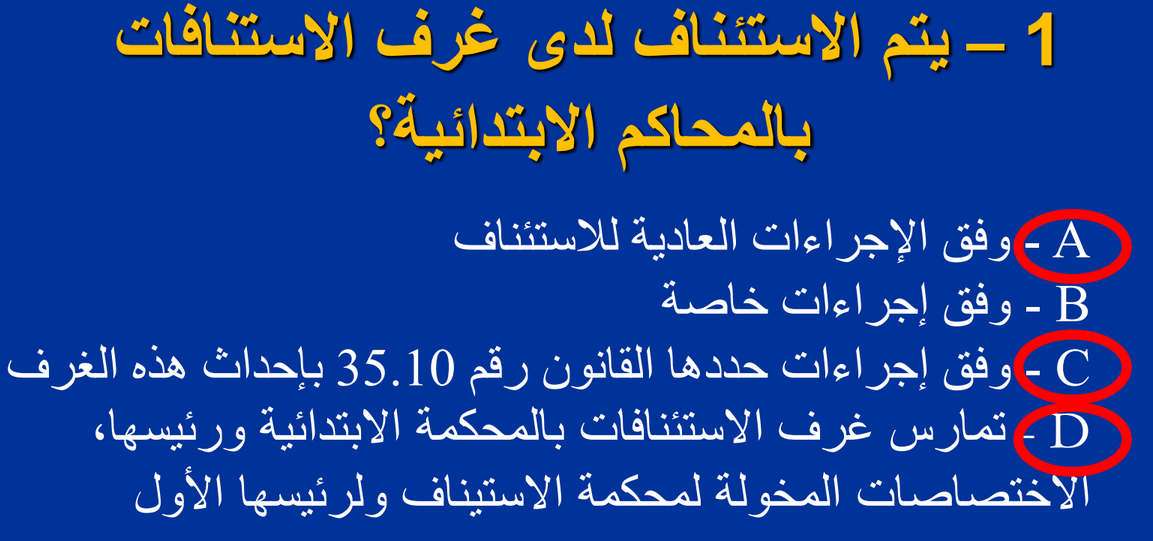}
        \caption{An example of a document page.}
        \label{fig:mcq_ex_1}
    \end{figure}

\begin{figure}[htbp]
        \centering
        \includegraphics[scale=0.6]{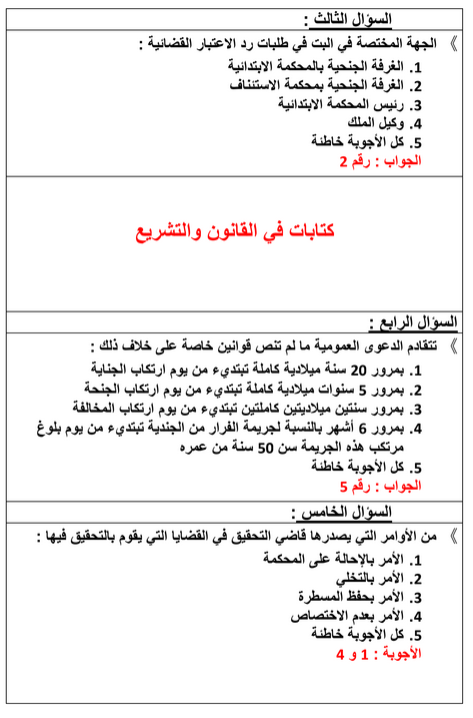}
        \caption{An example of a document page.}
        \label{fig:mcq_ex_2}
    \end{figure}

    The data is collected from a plethora of documents that are generally PDFs or Word documents. The MCQs are structured in various formats inside the documents: single MCQ per page (Figure \ref{fig:mcq_ex_1}), multiple MCQ per page (Figure \ref{fig:mcq_ex_2}), etc.
\subsection{Step 2: Temporal curation}
    The raw documents are given to a legal expert to evaluate the recency of the legislation that they are based on. The documents based on outdated legislation are not considered for further processing.
    
\subsection{Step 3: Organisation}
\begin{figure}
        \centering
        \includegraphics[scale=0.15]{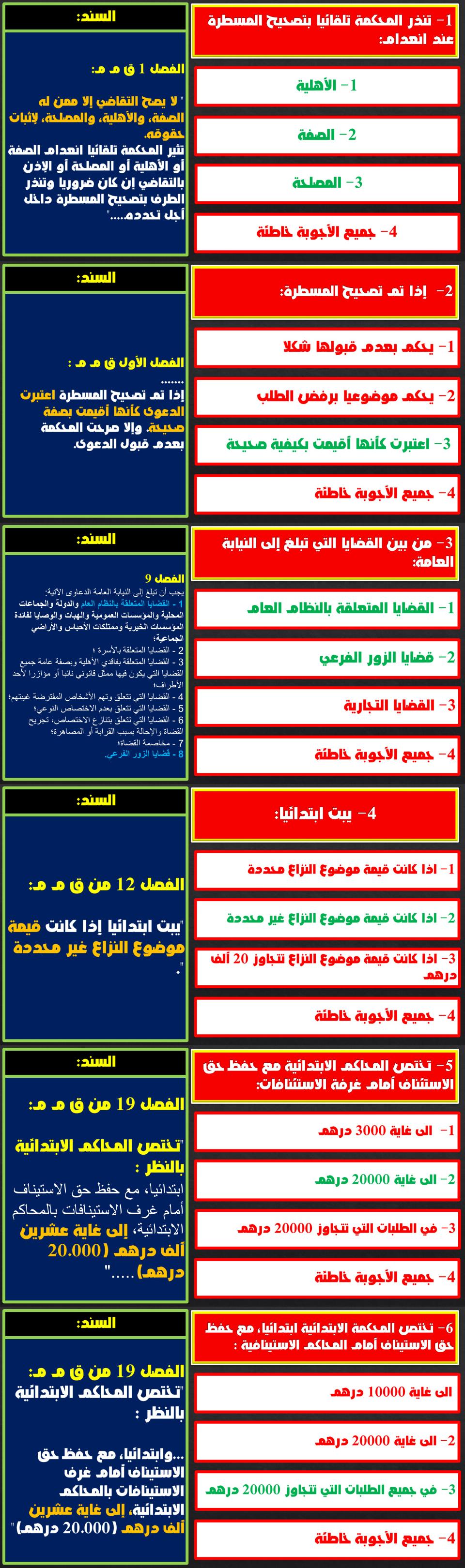}
        \caption{A batch of MCQs concatenated one below the other.}
        \label{fig:mcq_batch}
    \end{figure}

    The chosen documents are then either manually or automatically transformed into images containing batches of MCQs. The automatic process takes advantage of the structured nature of some documents to gather them in batches. On the other hand, for more irregular documents (e.g. pages contain a varying number of MCQs, MCQs that are not completely expressed on the same page, the answers for MCQs are in a separate page, etc). In this case, the MCQs are screened one by one manually and concatenated into a document. The pages of the documents are then turned into images. Figure \ref{fig:mcq_batch} shows an example of these images.
\subsection{Step 4: Extraction}

    After organising the MCQs to images, where each image contains a batch of MCQs, the images are fed to a vision LLM (Gemini-2.0.Flash) to structure the MCQs in a machine-readable format automatically. Figure \ref{fig:ex_prompt}represents the prompt used to extract MCQs.
    
    \begin{figure}[htbp]
        \centering
        \scalebox{0.8}{%
        \begin{tcolorbox}[colback=white, colframe=black, title=QA Pairs extration from images]
        \small
        \#\textbf{Instructions}:
        - This is a list of multiple-choice questions in Arabic.\\
        - Extract the different MCQS in the following format:\\
        
        [\\
            {\\
                "question": "",\\
                "options":{\\
                    "A": "",\\
                    "B": "",\\
                    "C": "",\\
                    "D": ""\\
                },\\
                "answer": "option index letter",\\
                "hint":"",\\
                "source": ""\\
            },\\
            ...\\
        ]\\
        
        \# \textbf{Response}:
        \end{tcolorbox}
        }
        \caption{Prompt for extracting MCQs from the organised images of MCQs obtained in step 3.}
        \label{fig:ex_prompt}
    \end{figure}
\subsection{Step 5: Verification}
\label{sec:verif_guid}
In this step, the MCQs are manually verified by annotators. The verification step follows the following guidelines:
\begin{itemize}
    \item Check if the question is similar to the original question.
    \item Check if the options are correct.
    \item check if the order of options is the same.
    \item check if the answers are similar to the original answers.
\end{itemize}
\subsection{Step 6: Categorisation}
    The annotators are tasked to use the original documents from which the MCQs are extracted to categorise the different law texts that they are based on (e.g. Criminal Law, Constitution, etc.). These categories are explored and normalised to remove any redundancy.
    
\section{Benchmarking}
    \label{sec:bench_app}
\begin{figure}[htbp]
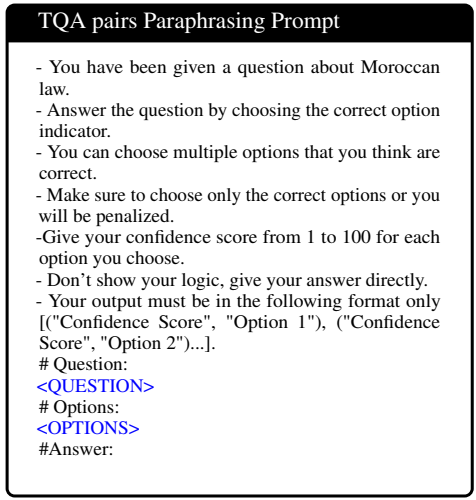

        \centering
        \scalebox{0.8}{%
        \begin{tcolorbox}[colback=white, colframe=black, title=TQA pairs Paraphrasing Prompt]
        \small
        - You have been given a question about Moroccan law.\\
        - Answer the question by choosing the correct option indicator.\\
        - You can choose multiple options that you think are correct.\\
        - Make sure to choose only the correct options or you will be penalized.\\
        -Give your confidence score from 1 to 100 for each option you choose.\\
        - Don't show your logic, give your answer directly.\\
        - Your output must be in the following format only [("Confidence Score", "Option 1"), ("Confidence Score", "Option 2")...].\\
        \# Question:\\
        \textcolor{blue}{<QUESTION>}\\
        \# Options:\\
        \textcolor{blue}{<OPTIONS>}\\
        \#Answer:\\
        \end{tcolorbox}
        }
        \caption{Englisg translation of instructions used to prompt various LLMs to answer MizanQA questions.}
        \label{fig:inference_prompt_eng_trans}
    \end{figure}

    MizanQA is tested on many multilingual and Arabic language models to assess their knowledge of Moroccan law. Figure \ref{fig:inference_prompt} shows the prompt for prompting the different LLMs. \ref{fig:inference_prompt_eng_trans} gives an english translation of the prompt.
    
    \begin{figure}[htbp]
        \centering
        \scalebox{0.8}{%
        \begin{tcolorbox}[colback=white, colframe=black, title=TQA pairs Paraphrasing Prompt]
        \small
        - \foreignlanguage{arabic}{لقد تم إعطاؤك سؤال حول القانون المغربي.}\\
        - \foreignlanguage{arabic}{أجب عن السؤال باختيار مؤشر الخيارات الصحيح.}\\
        - \foreignlanguage{arabic}{يمكنك اختيار خيارات متعددة تعتقد أنها صحيحة.}\\
        - \foreignlanguage{arabic}{تأكد من اختيار الخيارات الصحيحة فقط وإلا ستتعرض للعقوبة.}\\
        -\foreignlanguage{arabic}{أعطي درجة ثقتك من 1 إلى 100 في كل خيار تختاره.}\\
        - \foreignlanguage{arabic}{لا تظهر منطقك، بل أعط إجابتك مباشرة.}\\
        - \foreignlanguage{arabic}{جب أن يكون الناتج الخاص بك بالتنسيق التالي فقط لا غير [("درجة الثقة", "الخيار 1")، ("درجة الثقة", "الخيار 2") ...].}\\
        \# \foreignlanguage{arabic}{سؤال:}\\
        \textcolor{blue}{<QUESTION>}\\
        \# \foreignlanguage{arabic}{خيارات:}\\
        \textcolor{blue}{<OPTIONS>}\\
        \#\foreignlanguage{arabic}{ إجابة:}\\
        \end{tcolorbox}
        }
        \caption{Instructions used to prompt various LLMs to answer MizanQA questions.}
        \label{fig:inference_prompt}
    \end{figure}

    \subsection{Performance By Law Category}
    
    Table \ref{tab:res_table_by_cat} summarises the results of the different models by law category. The models are assessed across several Moroccan law categories: Civil Procedure, Criminal Law, Family Code, Family Law, Law of Obligations and Contracts, The Judicial System of the Kingdom, The Justice Sector, and The Moroccan Constitution. Across the models, there is a general trend of improvement in performance from Allam-2 (7b) to Gemini-2.0-flash, with the Gemini models generally outperforming the Llama models. For specific law categories, Law of Obligations and Contracts and The Moroccan Constitution tend to have higher scores across most metrics and models, indicating that these areas may be easier for the LLMs to handle. In addition to having affinities with laws from other countries (especially English-speaking). Conversely, Family Code and Criminal Law often exhibit lower performance scores, suggesting these domains pose a greater challenge. This can be attributed to the significant fusion between Islamic principles, the modern concepts of human rights adopted by Western countries. The calibration errors ($\text{ECE}_{\text{opt}}$ and $\text{ECE}_{\text{set}}$) vary across models and categories, with no clear pattern of consistency, indicating differences in the models' confidence and accuracy alignment.
    \begin{table*}[]
        \centering
        \setlength{\tabcolsep}{1.7pt}
        \scalebox{0.8}{
        \begin{tabular}{llrrrrrrr}
        \toprule
         Model & Category & PMPA(1) & PMPA(0.5) & F1-Like(1) & F1-Like(2) & ACC & $\text{ECE}_{\text{opt}}$ & $\text{ECE}_{\text{set}}$ \\
        \midrule
        \multirow[t]{8}{*}{Allam-2 (7b)} & Civil Procedure & 27.70 & 35.34 & 46.28 & 42.98 & 10.87 & 21.09 & 52.88 \\
         & Criminal Law & 26.73 & 32.90 & 40.94 & 38.00 & 17.95 & 33.02 & 50.62 \\
         & Family Code & 20.61 & 25.00 & 33.60 & 31.62 & 7.89 & 37.83 & 67.03 \\
         & Family Law & 31.69 & 39.71 & 50.13 & 47.33 & 13.64 & 27.36 & 56.62 \\
         & Law of Obligations and Contracts & 31.08 & 38.51 & 46.76 & 44.14 & 18.92 & 19.45 & 52.75 \\
         & The Judicial System of the Kingdom & 17.61 & 27.46 & 36.66 & 32.90 & 6.82 & 27.61 & 48.98 \\
         & The Justice Sector & 27.35 & 35.68 & 47.48 & 43.13 & 17.95 & 35.02 & 57.66 \\
         & The Moroccan Constitution & 41.67 & 54.64 & 64.40 & 59.69 & 28.57 & 24.62 & 40.50 \\
        \cline{1-9}
        \multirow[t]{8}{*}{Gemini-1.5-flash} & Civil Procedure & 40.50 & 50.66 & 61.79 & 56.72 & 25.85 & 19.02 & 43.48 \\
         & Criminal Law & 29.55 & 35.99 & 44.19 & 40.48 & 19.45 & 47.02 & 53.85 \\
         & Family Code & 48.68 & 54.61 & 63.51 & 59.52 & 34.21 & 22.21 & 51.46 \\
         & Family Law & 39.07 & 50.77 & 63.16 & 57.04 & 18.18 & 21.81 & 52.12 \\
         & Law of Obligations and Contracts & 70.27 & 79.05 & 84.41 & 80.77 & 62.16 & 14.94 & 22.15 \\
         & The Judicial System of the Kingdom & 39.32 & 47.44 & 54.07 & 50.06 & 29.49 & 29.44 & 47.19 \\
         & The Justice Sector & 42.31 & 55.56 & 62.54 & 56.49 & 30.77 & 26.31 & 50.79 \\
         & The Moroccan Constitution & 49.75 & 60.61 & 66.85 & 62.60 & 40.91 & 17.10 & 32.25 \\
        \cline{1-9}
        \multirow[t]{8}{*}{Gemini-2.0-flash} & Civil Procedure & 56.63 & 62.65 & 69.35 & 66.70 & 40.09 & 12.94 & 40.54 \\
         & Criminal Law & 48.37 & 51.86 & 58.72 & 56.04 & 39.55 & 40.25 & 44.13 \\
         & Family Code & 62.28 & 64.91 & 69.04 & 67.54 & 55.26 & 17.44 & 34.49 \\
         & Family Law & 60.23 & 66.91 & 73.51 & 70.72 & 40.91 & 13.13 & 41.31 \\
         & Law of Obligations and Contracts & 73.42 & 77.48 & 81.62 & 80.18 & 64.86 & 11.35 & 25.74 \\
         & The Judicial System of the Kingdom & 52.49 & 57.71 & 63.92 & 61.18 & 39.08 & 21.74 & 43.95 \\
         & The Justice Sector & 53.42 & 67.09 & 74.67 & 68.21 & 38.46 & 18.64 & 37.88 \\
         & The Moroccan Constitution & 69.76 & 75.12 & 80.29 & 78.00 & 58.57 & 11.43 & 30.61 \\
        \cline{1-9}
        \multirow[t]{8}{*}{Llama-3.3 (70b)} & Civil Procedure & 48.29 & 53.37 & 61.47 & 58.97 & 29.57 & 22.63 & 61.61 \\
         & Criminal Law & 44.24 & 47.38 & 57.52 & 53.79 & 33.29 & 44.85 & 60.60 \\
         & Family Code & 47.37 & 52.63 & 57.98 & 55.96 & 34.21 & 30.00 & 60.50 \\
         & Family Law & 42.75 & 49.43 & 56.20 & 53.21 & 21.21 & 25.82 & 66.62 \\
         & Law of Obligations and Contracts & 66.67 & 69.82 & 73.40 & 72.12 & 59.46 & 17.96 & 37.40 \\
         & The Judicial System of the Kingdom & 42.33 & 46.92 & 53.74 & 50.92 & 29.55 & 32.00 & 60.75 \\
         & The Justice Sector & 58.12 & 65.49 & 73.99 & 69.12 & 43.59 & 24.01 & 51.21 \\
         & The Moroccan Constitution & 59.05 & 62.14 & 67.46 & 65.98 & 45.71 & 17.88 & 48.85 \\
        \cline{1-9}
        \multirow[t]{8}{*}{Llama-4-maverick (17b)} & Civil Procedure & 53.86 & 59.98 & 67.61 & 65.17 & 35.15 & 7.55 & 33.10 \\
         & Criminal Law & 46.16 & 50.90 & 63.01 & 58.32 & 35.70 & 26.38 & 28.35 \\
         & Family Code & 56.14 & 60.75 & 65.18 & 63.33 & 42.11 & 9.84 & 30.08 \\
         & Family Law & 47.78 & 54.75 & 63.47 & 60.43 & 24.24 & 13.12 & 37.67 \\
         & Law of Obligations and Contracts & 72.97 & 78.38 & 82.52 & 80.36 & 64.86 & 5.92 & 26.88 \\
         & The Judicial System of the Kingdom & 46.31 & 53.03 & 59.78 & 56.70 & 34.09 & 13.48 & 37.28 \\
         & The Justice Sector & 51.92 & 61.11 & 68.64 & 63.93 & 41.03 & 9.72 & 23.45 \\
         & The Moroccan Constitution & 61.90 & 67.98 & 72.86 & 70.67 & 54.29 & 8.35 & 29.89 \\
        \cline{1-9}
        \multirow[t]{8}{*}{Llama-4-scout (17b)} & Civil Procedure & 52.26 & 57.20 & 64.96 & 62.62 & 34.78 & 22.09 & 57.10 \\
         & Criminal Law & 36.94 & 41.68 & 56.18 & 50.63 & 26.09 & 48.03 & 68.15 \\
         & Family Code & 50.00 & 55.26 & 60.18 & 58.25 & 39.47 & 29.33 & 56.42 \\
         & Family Law & 44.44 & 49.65 & 57.30 & 54.95 & 25.76 & 26.41 & 69.21 \\
         & Law of Obligations and Contracts & 69.82 & 74.32 & 78.17 & 76.17 & 59.46 & 16.62 & 35.33 \\
         & The Judicial System of the Kingdom & 38.92 & 43.37 & 49.47 & 47.30 & 26.14 & 32.22 & 61.12 \\
         & The Justice Sector & 44.66 & 56.20 & 67.20 & 60.67 & 33.33 & 31.06 & 55.47 \\
         & The Moroccan Constitution & 65.10 & 69.44 & 75.25 & 73.22 & 55.07 & 16.94 & 41.67 \\
        \cline{1-9}
        \bottomrule
        \end{tabular}}
        \caption{The results of different models on MizanQA, stratified by Moroccan law categories.}
        \label{tab:res_table_by_cat}
    \end{table*}
    \section{Technical setup}
        All the experiments are conducted using either the Groq API or the Gemini API. All the models are incorporated in Groq except Gemini-2.0-Flash and Gemini-1.5-Flash. We use Python to access the APIs, prompt the models, process and save their outputs.
    \section{Annotators}
        This dataset was annotated by volunteers. The group of volunteers contained one legal expert, three PhD students and one postdoctoral student, supervised by a professor. These participants agreed to volunteer for free due to the importance of the dataset in the assessment of legal knowledge in LLMs, which is a first step towards democratising access to legal support in Morocco. These annotators belong to a diverse set of demographic and socioeconomically backgrounds.
    \section{Use of AI}
        AI has been used in the extraction process. It was also evaluated using our dataset. During the writing of the paper, it was used for editing and grammar and style correction.
\end{document}